\documentclass[letterpaper]{article} 
\usepackage{aaai24}  
\usepackage{times}  
\usepackage{helvet}  
\usepackage{courier}  
\usepackage[hyphens]{url}  
\usepackage{graphicx} 
\usepackage{natbib}  
\usepackage{caption} 
\usepackage{algorithm}
\usepackage{algorithmic}
\usepackage{amsmath}
\usepackage{pifont}
\usepackage{multirow, booktabs}
\usepackage{amssymb}

\usepackage{newfloat}
\usepackage{listings}
\DeclareCaptionStyle{ruled}{labelfont=normalfont,labelsep=colon,strut=off} 
\floatstyle{ruled}
\newfloat{listing}{tb}{lst}{}
\floatname{listing}{Listing}
\title{BLADE: Box-Level Supervised Amodal Segmentation\\through Directed Expansion}
\author{
    Zhaochen Liu\textsuperscript{\rm 1,2}\equalcontrib,
    Zhixuan Li\textsuperscript{\rm 3}\equalcontrib,
    Tingting Jiang\textsuperscript{\rm 1,4}\thanks{Corresponding author.}
}
\affiliations{
    \textsuperscript{\rm 1}National Engineering Research Center of Visual Technology, National Key Laboratory \\
    for Multimedia Information Processing, School of Computer Science, Peking University\\
    \textsuperscript{\rm 2}AI Innovation Center, School of Computer Science, Peking University\\
    \textsuperscript{\rm 3}School of Computer Science and Engineering, Nanyang Technological University\\
    \textsuperscript{\rm 4}National Biomedical Imaging Center, Peking University\\

    \{dreamerliu, ttjiang\}@pku.edu.cn, zhixuanli520@gmail.com
}

\begin{document}

\maketitle

\begin{abstract}
Perceiving the complete shape of occluded objects is essential for human and machine intelligence. While the amodal segmentation task is to predict the complete mask of partially occluded objects, it is time-consuming and labor-intensive to annotate the pixel-level ground truth amodal masks. Box-level supervised amodal segmentation addresses this challenge by relying solely on ground truth bounding boxes and instance classes as supervision, thereby alleviating the need for exhaustive pixel-level annotations. Nevertheless, current box-level methodologies encounter limitations in generating low-resolution masks and imprecise boundaries, failing to meet the demands of practical real-world applications. We present a novel solution to tackle this problem by introducing a directed expansion approach from visible masks to corresponding amodal masks. Our approach involves a hybrid end-to-end network based on the overlapping region - the area where different instances intersect. Diverse segmentation strategies are applied for overlapping regions and non-overlapping regions according to distinct characteristics. To guide the expansion of visible masks, we introduce an elaborately-designed connectivity loss for overlapping regions, which leverages correlations with visible masks and facilitates accurate amodal segmentation. Experiments are conducted on several challenging datasets and the results show that our proposed method can outperform existing state-of-the-art methods with large margins.
\end{abstract}

\section{Introduction}
\emph{Amodal perception} is a vital ability of human's cognitive system~\cite{nanay2018importance, kanizsa1979organization} for inferring the complete shape of occluded objects easily. Amodal perception shows essential potential for tremendous real-world applications including autonomous driving~\cite{qi2019amodal, breitenstein2022amodal}, robotic gripping~\cite{wada2018instance, wada2019joint} and novel view synthesis~\cite{li20222d, gkitsas2021panodr}. For example, deducting the complete shape and range from the visible region of target objects (pedestrians or vehicles) is critical for accurate object recognition and routine planning in self-driving~\cite{ao2023image}.

\begin{figure}[t]
\centering
\includegraphics[width=0.8\columnwidth]{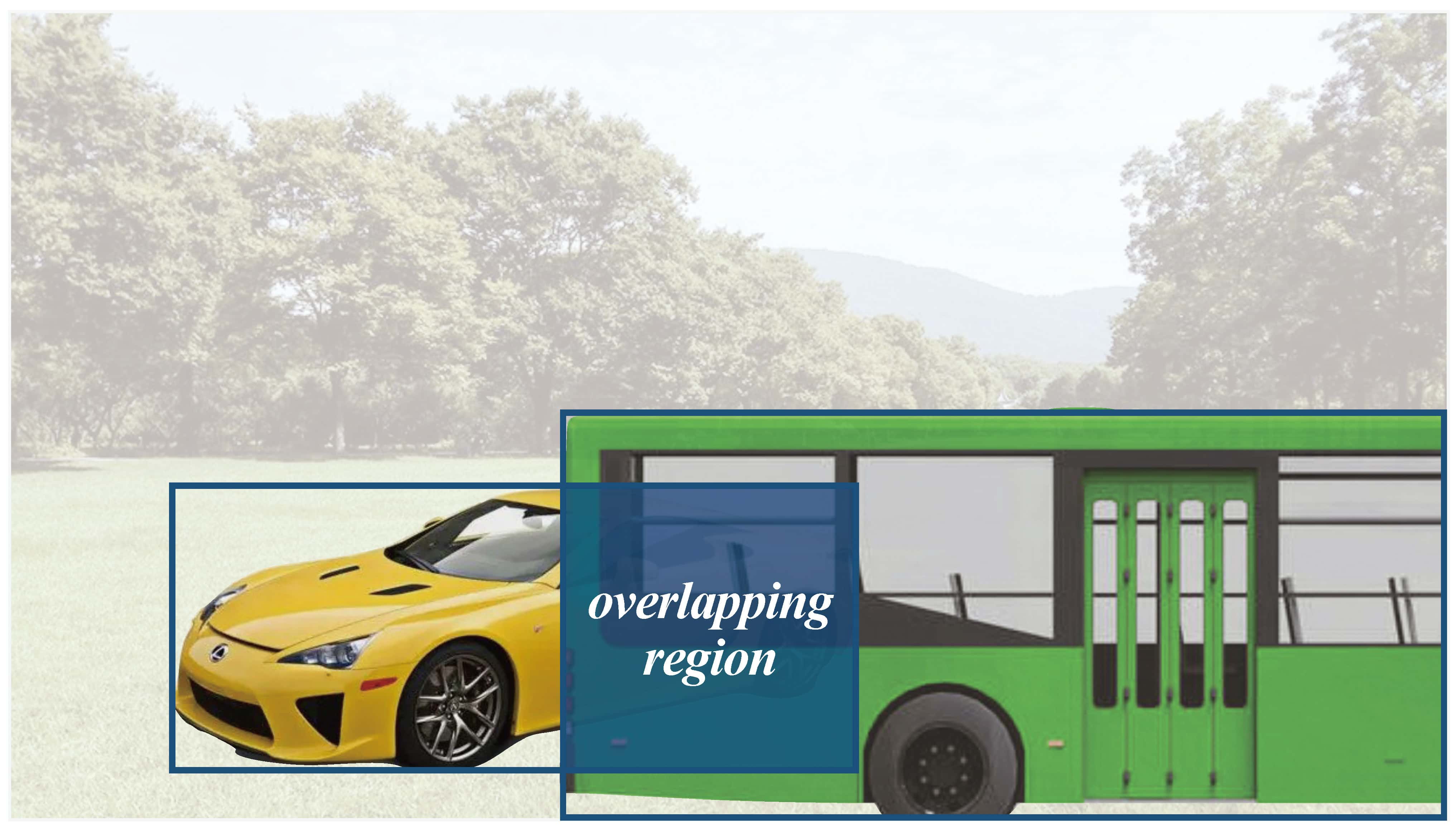}
\caption{An illustration of the overlapping region. The overlapping region of an object is the tightest bounding box that covers all intersecting areas of its amodal bounding box and those of other objects, so the occluded portion of each object should be inside if exists.}
\label{region1}
\end{figure}

In computer vision, amodal instance segmentation has aroused broad concern since it was proposed in AIS~\cite{li2016amodal}, which aims to predict complete shapes of partially occluded objects. However, annotating pixel-level ground-truth amodal masks for such objects is labor-intensive and error-prone due to the absence of visible cues in occluded regions. To mitigate the challenges of pixel-level annotation, Bayesian-Amodal~\cite{sun2022amodal}, a weakly supervised approach is proposed that utilizes ground-truth bounding boxes as an alternative supervision signal instead of the intricate ground-truth amodal masks. This method employs a Bayesian model to effectively address the amodal segmentation problem. Nevertheless, the amodal mask generated by the Bayesian-Amodal approach exhibits low resolution and uneven boundaries. This outcome arises from the Bayesian model's inherent coarse shape priors under box-level supervision, which inadequately align with the demands of real-world applications.

How to obtain amodal masks with both high-resolution and accurate boundaries solely through box-level supervision? To deal with this challenge, we propose the \textbf{B}ox-\textbf{L}evel supervised \textbf{A}modal segmentation network through \textbf{D}irected \textbf{E}xpansion~(\textbf{BLADE}), a weakly-supervised amodal segmentation method. The key insight of BLADE is to first deduct the visible mask from the detected bounding box, and then expand it to the amodal mask with the guidance of the correlation between the two masks. The correlation reflects the resemblance and distinction between visible and occluded regions in terms of shape and appearance, thus indicating the direction and extent of the expansion. Compared with expanding in a naively unguided manner, the proposed approach can provide more explicit guidance, easing the network's learning burden and contributing to more accurate amodal mask prediction. To depict the aforementioned correlation, a new hybrid end-to-end network is proposed based on the \textit{overlapping region}. As shown in Fig.~\ref{region1}, the overlapping region reveals the intersection between an instance and others, encompassing the occluded portion of the instance. And expansion-encouraged and relatively conservative strategies are designed for overlapping regions and non-overlapping regions, respectively.

Specifically, our method extends the box-level supervised instance segmentation technique introduced in BoxInst~\cite{tian2021boxinst}. While BoxInst effectively performs segmentation on non-overlapping regions, we broaden its capabilities to enable amodal segmentation. Our proposed network first predicts the visible mask and corresponding expanded coarse amodal mask, and then fuses them according to the estimated overlapping region. To achieve this, besides the original visible branch contained by BoxInst, we introduce two additional branches, namely amodal-branch and region-branch, to distinguish overlapping regions and non-overlapping regions. The three branches, visible-branch, amodal-branch, and region-branch, predict the visible mask, the coarse amodal mask, and the position of overlapping region, respectively. As for the amodal-branch, observing that within the overlapping region, the visible portion of an object maintains a significant adjacency with its occluded counterpart, we design a connectivity loss to encourage the expansion of the visible segment within the overlapping area towards its encompassing vicinity, finally reaching the coverage of occluded segments through the cooperation with other losses. As for the visible-branch, a general segmentation approach is adopted in which there is no expansion-encouraged factor. The final output amodal prediction is determined by combining the outcomes of three branches. Concretely, for the overlapping region, the segmentation result is from the amodal-branch, while in the remaining areas the segmentation result is from the visible-branch.

We have conducted experiments on three challenging datasets, including OccludedVehicles~\cite{wang2020robust}, KINS~\cite{qi2019amodal} and COCOA-cls~\cite{follmann2019learning}. The results show that our proposed approach outperforms existing weakly-supervised methods with large margins and significantly reduces the performance gap with fully-supervised methods. Our contributions can be summarized as follows:

\begin{itemize}
\item We introduce a novel hybrid end-to-end network utilizing the overlapping region. This approach enables diverse segments of an instance to employ tailored segmentation strategies while facilitating collaborative interaction.
\item We propose a novel connectivity loss for the overlapping region, guiding the visible mask to expand towards the amodal mask. This approach leverages the correlation with the visible segment, facilitating the accurate prediction of occluded components.
\item Our approach significantly outperforms the existing box-level supervised instance segmentation method, reaching state-of-the-art performance.
\end{itemize}

\begin{figure*}[ht]
\centering
\includegraphics[width=2\columnwidth]{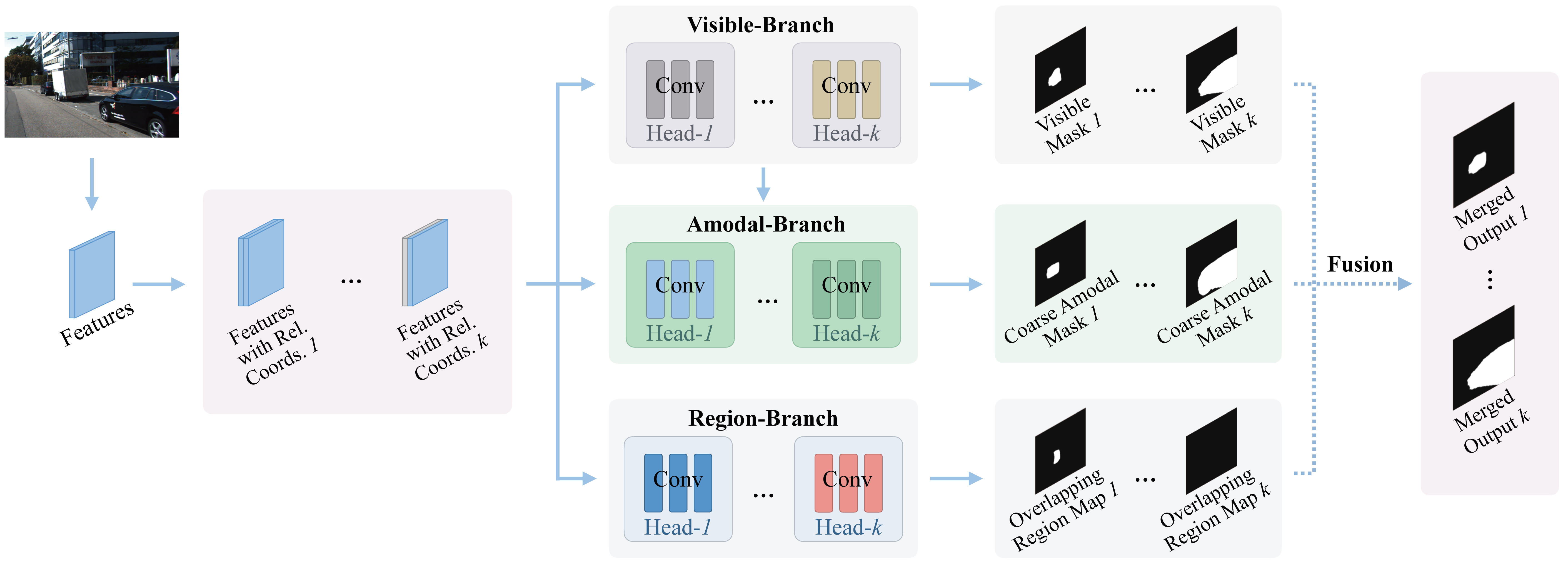}
\caption{A schematic illustration of the proposed BLADE approach. Extracted features with relative coordinates maps are input to visible-branch, amodal-branch, and region-branch to predict the visible mask, coarse amodal mask, and overlapping region map of each instance respectively, which all adopt dynamically generated instance-aware mask heads. Exploiting the correlation, predicted visible masks are also input to the amodal branch for our proposed connectivity loss that directs the expansion from visible masks to corresponding coarse amodal masks. The final outputs use coarse amodal masks in predicted overlapping regions and visible masks in other regions.}
\label{overview}
\end{figure*}

\section{Related Work}

\textbf{Amodal segmentation} is to predict the shape of both visible parts and occluded parts of partially occluded objects. Malik first proposed the task and provided a synthetic dataset~\cite{li2016amodal}. Subsequently, the KINS~\cite{qi2019amodal} and Amodal COCO~\cite{zhu2017semantic} datasets based on real-world images and human-predicted annotations were built. Many fully supervised approaches have been proposed in this field. In addition to direct optimization methods~\cite{li2016amodal, zhu2017semantic, qi2019amodal}, researchers have introduced and utilized depth relationships~\cite{zhang2019learning}, shape priors~\cite{xiao2021amodal, li20222d}, compositional models~\cite{wang2020robust} and the correlation between visible and occluded segments~\cite{follmann2019learning, ke2021deep} to improve performance. And some related work~\cite{ehsani2018segan, dhamo2019object, ling2020variational} achieves amodal segmentation based on amodal perception and completion. However, fully supervised methods face a common problem that data annotation takes considerable time and effort. Especially for real-world images, humans can only figure out the approximate shape and range of the occluded parts of an object based on experience, which may vary due to different annotators and increase system error. Noticing this problem, some researchers proposed self-supervised amodal perception and completion methods~\cite{zhan2020self, nguyen2021weakly}. Nevertheless, these methods introduce the correlation with occluders, requiring the category of occluders during training and testing, which brings great limitations to practical applications. In contrast, only bounding boxes and class annotations of instances are adopted as supervision in our approach thus avoiding the above issues.

\noindent \textbf{Box-level supervised amodal segmentation} was proposed in Bayesian-Amodal~\cite{sun2022amodal}, which is exactly the setting adopted by our method. Box-level supervised amodal segmentation uses bounding boxes and categories of instances as supervision during training, which tackles time-consuming, labor-intensive, and error-prone pixel-level labeling and better supports larger-scale training. Based on related work~\cite{kortylewski2020compositional, kortylewski2021compositional}, Bayesian-Amodal also proposed a method that outperformed alternative weakly supervised methods, which replaced the fully connected classifier in neural networks with a Bayesian generative model of the neural network features. However, the results of this method have a relatively low resolution so we cannot get more detailed information. In this paper, we propose a new hybrid network with better accuracy, in which distinct and proper strategies are adopted for different regions and the correlation between visible and occluded regions is utilized.

\noindent \textbf{Box-level supervised segmentation} is an important and concerned field of computer vision, which comparatively focuses on the segmentation of non-occluded instances. Box-level supervision is weak, so related methods often introduce some observations and priors as assistance. SDI~\cite{khoreva2017simple} utilizes the object shape priors. OSIS~\cite{pham2018bayesian} introduces the Bayesian model for better formulation. BBTP~\cite{hsu2019weakly} exploits the bounding box tightness prior. WSIS~\cite{arun2020weakly} builds an annotation consistency framework. In~\cite{sun2020weakly}, a two-stage transfer learning framework employing valid generated masks from GrabCut~\cite{rother2004grabcut} is designed. Recently, BoxInst~\cite{tian2021boxinst} raises an observation that proximal pixels with similar colors tend to have the same label, and proposes a concise and effective method incorporating a projection loss and a pairwise loss. BoxInst achieves leading performance and greatly narrows the performance gap between weakly and fully supervised instance segmentation. However, these methods do not perform well if directly transferred to an amodal segmentation task. In our approach, a novel connectivity loss encourages the mask to expand from the visible region towards the occluded region is designed and helps achieve the goal of amodal segmentation.

\section{Methodology Design}
\subsection{Problem Definition}
Amodal segmentation aims to predict the amodal mask $\mathbf{M}_a$ for a given image $\mathbf{I}$ and some region-of-interest~\cite{xiao2021amodal}. In this paper, we follow the task setting of box-level supervised amodal segmentation and take only bounding box annotations and class labels as the supervision signal. 

Specifically, we take ground-truth $\mathbf{B}_v$~(the bounding box of visible portion), $\mathbf{B}_a$~(the bounding box of the complete object), and $\mathbf{c}$~(the class label) of each instance as the supervision for training. During the test, we take only $\mathbf{B}_{test}$, the object bounding box of the visible area, as input to determine the region of interest and match the corresponding instance. The network ultimately outputs an amodal mask $\mathbf{m}$ of the instance.

\subsection{Network Architecture}
Apparently, $\mathbf{M}_a$ can be decomposed as
\begin{equation}
    \mathbf{M}_a = \mathbf{M}_v + \mathbf{M}_o,
\end{equation}
where $\mathbf{M}_v$ and $\mathbf{M}_o$ represent the visible mask and the occluded mask respectively. The prediction of $\mathbf{M}_v$ can exactly be formulated as a less complicated general segmentation problem and provide the estimation of $\mathbf{M}_o$ with clues. 

Inspired by this, we design a hybrid structure with multiple branches: as shown in Fig.~\ref{overview}, our network consists of visible-branch, amodal-branch, and region-branch, which are used to predict the visible mask $\mathbf{m}_v$, the coarse amodal mask $\mathbf{m}_a$, and the map of the overlapping region $\mathbf{m}_r$ respectively. The final output $\mathbf{m}$ fuses these three items, using $\mathbf{m}_a$ in the predicted overlapping region while using $\mathbf{m}_v$ in the other region. That is to say
\begin{equation}
    \mathbf{m} = \mathbf{m}_a * \mathbf{m}_r + \mathbf{m}_v * \mathbf{\bar{m}}_r,
\end{equation}
where $\mathbf{\bar{m}}_r$ is inverse $\mathbf{m}_r$ indicates non-overlapping region.

Our proposed network is developed on the recent box-level supervised instance segmentation method BoxInst~\cite{tian2021boxinst}. For the visible-branch, $\mathbf{B}_v$ annotations are applied as the supervision. We directly use the original mask heads with projection loss and pairwise loss in BoxInst which are set up just for our demand. For the amodal-branch, $\mathbf{B}_a$ annotations are applied as the supervision. We feed the amodal-branch the predicted $\mathbf{m}_v$ from visible-branch in addition to the features and relative coordinates. Utilizing the input $\mathbf{m}_v$ as clues, we introduce a connectivity loss to direct the expansion from $\mathbf{m}_v$ to $\mathbf{m}_a$. Meanwhile, projection loss and pairwise loss are also used to limit the range of expansion. For the region-branch, we transform the prediction of the four parameters of some ground truth bounding box into the prediction of the corresponding bitmask to improve robustness while using a simple pixel-level BCE loss. The above three branches share the same multi-scale features extracted from the image and all adopt dynamically-generated instance-aware mask heads containing varying instance-by-instance parameters~\cite{jia2016dynamic}, thereby significantly increasing the flexibility and decreasing the amount of parameters. The entire network is trained together, so the multiple branches form joint supervision, in which the correlation between the visible region and the occluded region is implied and mutual assistance is built.

\begin{figure}[t]
\centering
\includegraphics[width=0.71\columnwidth]{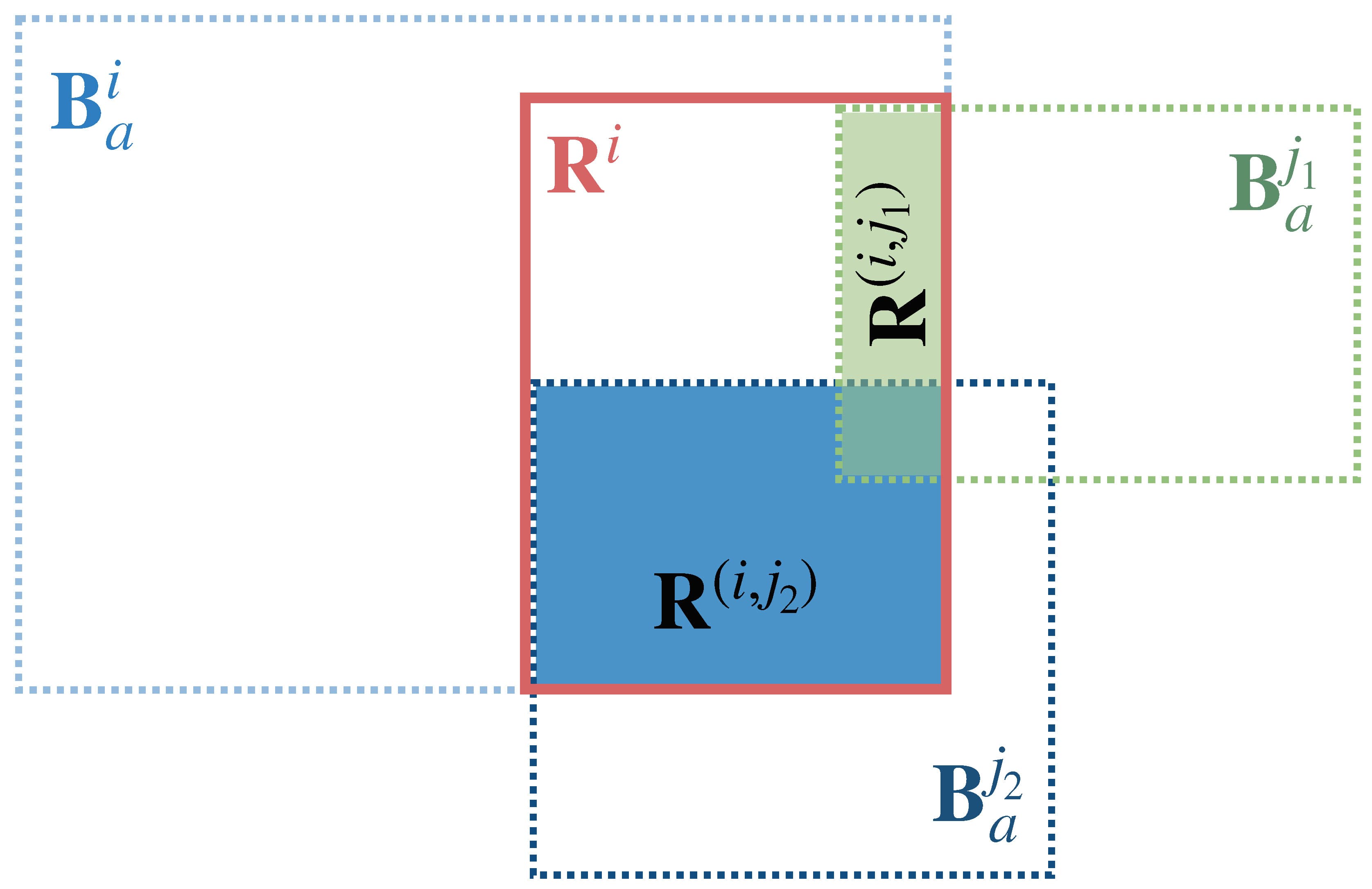}
\caption{If there are multiple intersecting areas, the envelope box is used as the ground-truth overlapping region. For the example in the figure, both $\mathbf{B}_a^{j_1}$ and $\mathbf{B}_a^{j_2}$ overlaps $\mathbf{B}_a^{i}$, then the red box $\mathbf{R}^i$ is defined as the overlapping region of instance $i$.}
\label{region2}
\end{figure}

\subsection{Overlapping Region}
As shown in Fig.~\ref{region1}, the overlapping region is the area where different instances overlap. Overlapping and occlusion are two accompanying phenomena: if an object is occluded, it must overlap another object, which means the overlapping region contains the occluded region. Therefore, We can use the easily accessible overlapping region as an approximate estimation of the corresponding occluded region to determine where the expanded amodal mask $\mathbf{m}_a$ is applied.

Under the setting of box-level supervision, the overlapping region of each instance is recorded by one bounding box. We use the intersection between the amodal bounding box of each instance and those of other instances within the same image, obviating the necessity for introducing additional annotations. Suppose there are $n$ instances in an image, and their amodal bounding boxes are
\begin{equation}
    \mathbf{B}_a^i = (x_{min}^i, y_{min}^i, x_{max}^i, y_{max}^i), ~i=1,...,n,
\end{equation}
then for instance $i$, the overlapping region with instance $j$ is
\begin{equation}
    \mathbf{R}^{(i,j)}=(x_{min}^{(i,j)}, y_{min}^{(i,j)}, x_{max}^{(i,j)}, y_{max}^{(i,j)})
\end{equation}
if exists, where
\begin{align}
    &x_{min}^{(i,j)} = \max(x_{min}^i, x_{min}^j), ~y_{min}^{(i,j)} = \max(y_{min}^i, y_{min}^j), \notag \\ 
    &x_{max}^{(i,j)} = \min(x_{max}^i, x_{max}^j), ~y_{max}^{(i,j)}) = \min(y_{max}^i, y_{max}^j).
\end{align}
As shown in Fig.~\ref{region2}, we take the envelope box of all existing overlapping region $\mathbf{R}^{(i,j)}$ as the ground-truth overlapping region of instance $i$, which is
\begin{align}
    \mathbf{R}^i=(&\min_{j\in \mathbf{V}_i}\{x_{min}^{(i,j)}\}, \min_{j\in \mathbf{V}_i}\{y_{min}^{(i,j)}\}, \notag \\ 
    &\max_{j\in \mathbf{V}_i}\{x_{max}^{(i,j)}\}, \max_{j\in \mathbf{V}_i}\{y_{max}^{(i,j)}\}), 
\end{align}
where $\mathbf{V}_i$ is the indexes of all instances that possess the valid overlapping region with instance $i$. During training and testing, we actually use the corresponding bitmask instead of the four-parameter representation, in which the pixel value is $1$ if within the overlapping region and is $0$ if outside. A pixel-level BCE loss
\begin{equation}
    L^r = -\frac{\alpha^r}{N}\sum_{\mathbf{m}_r}p\log\Tilde{p}+(1-p)\log(1-\Tilde{p})
\end{equation}
is applied for region-branch accordingly, where $p,\Tilde{p}$ is the ground truth value and the predicted value of some pixel, $N$ is the total number of pixels and $\alpha^r$ is a constant coefficient.

\begin{figure}[t]
\centering
\includegraphics[width=0.88\columnwidth]{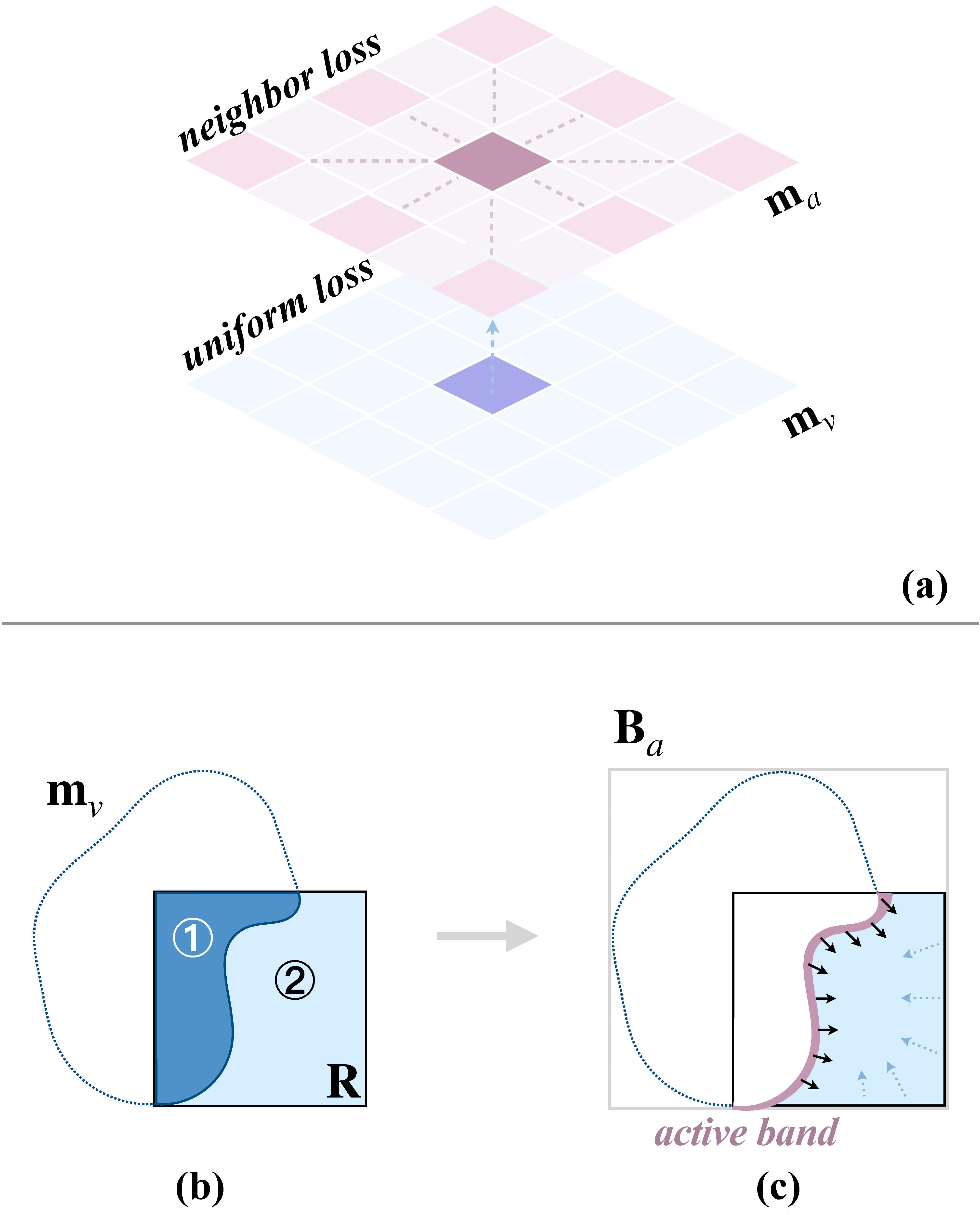}
\caption{An illustration of the connectivity loss. (a) The connectivity loss contains two terms, namely neighbor loss and uniform loss. The neighbor loss measures the label consistency of each pixel with its neighbors in $\mathbf{m}_a$, while the uniform loss reflects the consistency of corresponding pixels between $\mathbf{m}_a$ and $\mathbf{m}_v$. (b) The neighbor loss is applied to predicted-overlapping-visible pixels~(region \ding{172}), while the uniform loss is applied to the whole overlapping region $\mathbf{R}$~(region \ding{172}+\ding{173}). (c) By the action of the connectivity loss, an active band is built as the initiation of expansion. Multiple losses for the amodal-branch reach a balance of encouragement and inhibition of expansion thus directing a moderate expansion.}
\label{loss}
\end{figure}

\subsection{Connectivity Loss}
As mentioned above, the overall loss function of amodal-branch can be written as
\begin{equation}
    L^{a} = \alpha_1^a L_{proj}^a + \alpha_2^a L_{pair}^a + \alpha_3^a L_{con},
\end{equation}
where $\alpha_i^a$ is the constant weight of each term. As shown in Fig.~\ref{loss}, the newly-introduced connectivity loss consists of two parts
\begin{equation}
    L_{con} = l_{ne} + l_{un},
\end{equation}
namely neighbor loss and uniform loss.

The neighbor loss is applied to predicted-overlapping-visible pixels~(pixels in the overlapping region that are predicted to belong to the visible mask $\mathbf{m}_v$ in visible-branch) in the predicted coarse amodal mask $\mathbf{m}_a$, which measures the label consistency of each pixel with its neighbors. We design this loss based on an observation that in the overlapping region, the visible segment of an instance is mostly adjacent to its occluded counterpart. Consider an undirected graph $G=(V_{pov}, E_{pov})$. $V_{pov}$ is the set of predicted-overlapping-visible pixels satisfies
\begin{equation}
    \forall (i,j)\in V_{pov}, (i,j)\in\mathbf{R}~\land~\mathbf{m}_v(i,j)>t,
\end{equation}
where $\mathbf{m}_a(i,j)$ is the value of predicted coarse amodal mask $\mathbf{m}_a$ at position $(i,j)$, $t$ is the threshold of the visible-branch, and $E_{pov}$ is the set of edges that connect each pixel with its eight neighbors and contain at least one pixel in $V_{pov}$. For an edge $e=((i_1,j_1),(i_2,j_2))\in E_{pov}$, the ground-truth consistency value $c_e = 1$ when the labels of its two endpoints are the same 
\begin{equation}
    (i_1,j_1),(i_2,j_2)\in\mathbf{B}_a \vee (i_1,j_1),(i_2,j_2)\notin\mathbf{B}_a,
\end{equation}
while $c_e = 0$ when the labels are different. The predicted consistency value $\Tilde{c}_e$ can be defined as 
\begin{align}
    \Tilde{c_e} = &\mathbf{m}_a(i_1,j_1) ~\cdot~ \mathbf{m}_a(i_2,j_2) ~+ \notag \\
    &(1 - \mathbf{m}_a(i_1,j_1)) ~\cdot~ (1 - \mathbf{m}_a(i_2,j_2)),
\end{align}
which is positively correlated with the consistency and certainty. We adopt the frequently-used BCE loss 
\begin{equation}
    l_{ne} = -\frac{1}{N_e} \sum_{e\in E_{pov}} c_e\log\Tilde{c}_e + (1-c_e)\log(1-\Tilde{c}_e)
\end{equation}
to minimize the gap between all $\Tilde{c}_e$ and corresponding $c_e$, where $N_e$ is the number of edges in $E_{pov}$.

The uniform loss is applied to the whole overlapping region, which ensures the consistency between the predicted coarse amodal mask $\mathbf{m}_a$ and the predicted visible mask $\mathbf{m}_v$. The values $\mathbf{m}_a(i,j)$ and $\mathbf{m}_v(i,j)$ can be regarded as predictions of the probability that pixel $(i,j)$ belongs to the object and the probability it belongs to the visible portion of the object, respectively. Therefore, it's obvious that any $\mathbf{m}_a(i,j)$ should NOT be less than $\mathbf{m}_v(i,j)$. Observing this, the uniform loss is defined as 
\begin{equation}
    l_{un} = \frac{K}{N_\mathbf{R}} \sum_{(i,j)\in\mathbf{R}} \max(\mathbf{m}_v(i,j)-\mathbf{m}_a(i,j), 0)
\end{equation}
to penalize those pixels with reduced values from $\mathbf{m}_v$ to $\mathbf{m}_a$, where $K$ is a constant coefficient used to regulate the severity of the penalty, $\mathbf{R}$ is the set of pixels in the overlapping region and $N_\mathbf{R}$ is the number of these pixels.

By introducing the connectivity loss, we build an active band in the predicted coarse amodal mask $\mathbf{m}_a$ containing the neighbors of predicted-overlapping-visible pixels near edges, which tends to increase the value to share the same label as pixels in $\mathbf{m}_v$. We use neighbors with a one-pixel gap to the center pixel to increase the width of the active band, which can be adjusted as needed. The active band will further expand outward within a certain range until reaching a balanced state due to the existence of pairwise loss and projection loss. Intuitively, diverse losses adopted in amodal-branch form an antagonistic effect of encouragement and inhibition of expansion, resulting in both under-expansion and over-expansion being largely avoided.

\section{Experiments}

\subsection{Datasets and Metric}
\paragraph{Datasets.} Our experiments are conducted on three challenging datasets for amodal segmentation, including OccludedVehicles~\cite{wang2020robust}, KINS~\cite{qi2019amodal} and COCOA-cls~\cite{follmann2019learning}. 

The OccludedVehicles dataset is a synthetic dataset based on PASCAL3D+~\cite{xiang2014beyond} in which occluders are pasted randomly. It consists of 51801 objects that are evenly distributed among three foreground occlusion levels, FG-1, FG-2, FG-3, three background occlusion levels BG-1, BG-2, BG-3, and a non-occluded level FG-0. The three foreground occlusion levels correspond to 20-40\%, 40-60\%, and 60-80\% of the object being occluded, while the three background occlusion levels correspond to 1-20\%, 20-40\%, and 40-60\% of the context being occluded. 

The KINS dataset is based on real-world images with real occlusion. We follow the experimental settings of Bayesian-Amodal~\cite{sun2022amodal} that restricts the scope of evaluation to vehicles with a height greater than 50 pixels and divides the objects into four occlusion levels FG-0, FG-1, FG-2, FG-3, which correspond to 0\%, 1-30\%, 30-60\%, and 60-90\% of the object being occluded. The subset we adopt consists of 14826 objects. 

The COCOA-cls dataset is an extension of the real occlusion dataset Amodal COCO~\cite{zhu2017semantic} and consists of 766 objects. It's divided into four foreground occlusion levels FG-0, FG-1, FG-2, and FG-3, which correspond to 0\%, 1-20\%, 20-40\%, and 40-70\% of the object being occluded.

\paragraph{Metric.} As for the metric of evaluation, we adopt mean Intersection-over-Union~(IoU) like related work~\cite{sun2022amodal}. IoU calculates the ratio of intersecting pixels between the predicted amodal mask and corresponding ground truth amodal mask to their union, therefore a larger value indicates a more accurate segmentation result.

\begin{table*}[ht]
\centering
\resizebox{0.93\textwidth}{!}{%
\begin{tabular}{ccccccccccccc}
\toprule[1.4pt]
\multicolumn{1}{c|}{\multirow{3}{*}{Method}} &
\multicolumn{1}{c|}{\multirow{3}{*}{known $c$}} &
\multicolumn{11}{c}{OccludedVehicles} \\ \cline{3-13}
\multicolumn{1}{c|}{} &
  \multicolumn{1}{c|}{} &
  \multicolumn{1}{c|}{FG-0} &
  \multicolumn{3}{c|}{FG-1} &
  \multicolumn{3}{c|}{FG-2} &
  \multicolumn{3}{c|}{FG-3} &
  \multirow{2}{*}{Mean} \\ \cline{3-12}
\multicolumn{1}{c|}{} &
  \multicolumn{1}{c|}{} &
  \multicolumn{1}{c|}{-} &
  \multicolumn{1}{c|}{BG-1} &
  \multicolumn{1}{c|}{BG-2} &
  \multicolumn{1}{c|}{BG-3} &
  \multicolumn{1}{c|}{BG-1} &
  \multicolumn{1}{c|}{BG-2} &
  \multicolumn{1}{c|}{BG-3} &
  \multicolumn{1}{c|}{BG-1} &
  \multicolumn{1}{c|}{BG-2} &
  \multicolumn{1}{c|}{BG-3} &
   \\ \hline
\multicolumn{1}{c|}{BBTP} &
  \multicolumn{1}{c|}{Yes} &
  \multicolumn{1}{c|}{66.5} &
  \multicolumn{1}{c|}{59.7} &
  \multicolumn{1}{c|}{58.4} &
  \multicolumn{1}{c|}{57.9} &
  \multicolumn{1}{c|}{54.4} &
  \multicolumn{1}{c|}{51.0} &
  \multicolumn{1}{c|}{48.9} &
  \multicolumn{1}{c|}{50.4} &
  \multicolumn{1}{c|}{44.7} &
  \multicolumn{1}{c|}{40.2} &
  53.2 \\
\multicolumn{1}{c|}{BoxInst} &
  \multicolumn{1}{c|}{Yes} &
  \multicolumn{1}{c|}{72.3} &
  \multicolumn{1}{c|}{52.5} &
  \multicolumn{1}{c|}{53.5} &
  \multicolumn{1}{c|}{53.9} &
  \multicolumn{1}{c|}{37.7} &
  \multicolumn{1}{c|}{38.1} &
  \multicolumn{1}{c|}{38.2} &
  \multicolumn{1}{c|}{23.0} &
  \multicolumn{1}{c|}{22.8} &
  \multicolumn{1}{c|}{23.7} &
   41.6 \\
\multicolumn{1}{c|}{Bayesian-Amodal} &
  \multicolumn{1}{c|}{Yes} &
  \multicolumn{1}{c|}{63.9} &
  \multicolumn{1}{c|}{59.7} &
  \multicolumn{1}{c|}{59.6} &
  \multicolumn{1}{c|}{59.7} &
  \multicolumn{1}{c|}{57.2} &
  \multicolumn{1}{c|}{56.8} &
  \multicolumn{1}{c|}{56.8} &
  \multicolumn{1}{c|}{55.0} &
  \multicolumn{1}{c|}{53.9} &
  \multicolumn{1}{c|}{53.4} &
  57.6 \\
\multicolumn{1}{c|}{Bayesian-Amodal} &
  \multicolumn{1}{c|}{No} &
  \multicolumn{1}{c|}{63.0} &
  \multicolumn{1}{c|}{59.5} &
  \multicolumn{1}{c|}{59.5} &
  \multicolumn{1}{c|}{59.5} &
  \multicolumn{1}{c|}{56.2} &
  \multicolumn{1}{c|}{55.9} &
  \multicolumn{1}{c|}{55.6} &
  \multicolumn{1}{c|}{51.9} &
  \multicolumn{1}{c|}{50.6} &
  \multicolumn{1}{c|}{48.3} &
  56.0 \\ \hline
\multicolumn{1}{c|}{Ours} &
  \multicolumn{1}{c|}{No} &
  \multicolumn{1}{c|}{\textbf{73.2}} &
  \multicolumn{1}{c|}{\textbf{70.5}} &
  \multicolumn{1}{c|}{\textbf{69.7}} &
  \multicolumn{1}{c|}{\textbf{68.9}} &
  \multicolumn{1}{c|}{\textbf{69.7}} &
  \multicolumn{1}{c|}{\textbf{68.1}} &
  \multicolumn{1}{c|}{\textbf{66.2}} &
  \multicolumn{1}{c|}{\textbf{68.2}} &
  \multicolumn{1}{c|}{\textbf{64.5}} &
  \multicolumn{1}{c|}{\textbf{62.8}} &
  \textbf{68.2} \\ 
\end{tabular}%
}
\resizebox{0.96\textwidth}{!}{%
\begin{tabular}{c|c|c|ccccc|ccccc}
\toprule[1.4pt]
\multirow{2}{*}{Method} &
  \multirow{2}{*}{Supervision} &
  \multirow{2}{*}{known $c$} &
  \multicolumn{5}{c|}{KINS} &
  \multicolumn{5}{c}{COCOA-cls} \\ 
  \cline{4-13} 
 &
   &
   &
  \multicolumn{1}{c|}{FG-0} &
  \multicolumn{1}{c|}{FG-1} &
  \multicolumn{1}{c|}{FG-2} &
  \multicolumn{1}{c|}{FG-3} &
  Mean &
  \multicolumn{1}{c|}{FG-0} &
  \multicolumn{1}{c|}{FG-1} &
  \multicolumn{1}{c|}{FG-2} &
  \multicolumn{1}{c|}{FG-3} &
  Mean \\ \hline
SAM~(ViT-H) &
  - &
  Yes &
  \multicolumn{1}{c|}{86.7} &
  \multicolumn{1}{c|}{75.0} &
  \multicolumn{1}{c|}{50.8} &
  \multicolumn{1}{c|}{39.0} &
    62.9 &
  \multicolumn{1}{c|}{82.7} &
  \multicolumn{1}{c|}{74.9} &
  \multicolumn{1}{c|}{59.2} &
  \multicolumn{1}{c|}{42.3} &
   64.8 \\
VRSP &
  fully &
  - &
  \multicolumn{1}{c|}{84.7} &
  \multicolumn{1}{c|}{75.8} &
  \multicolumn{1}{c|}{74.5} &
  \multicolumn{1}{c|}{67.1} &
  75.5 &
  \multicolumn{1}{c|}{82.1} &
  \multicolumn{1}{c|}{77.7} &
  \multicolumn{1}{c|}{74.5} &
  \multicolumn{1}{c|}{72.9} &
  76.8 \\
AISFormer &
  fully &
  - &
  \multicolumn{1}{c|}{85.8} &
  \multicolumn{1}{c|}{76.4} &
  \multicolumn{1}{c|}{75.0} &
  \multicolumn{1}{c|}{69.4} &
  76.7 &
  \multicolumn{1}{c|}{80.6} &
  \multicolumn{1}{c|}{76.9} &
  \multicolumn{1}{c|}{70.9} &
  \multicolumn{1}{c|}{62.1} &
  72.6 \\ \hline
BBTP &
  weakly &
  Yes &
  \multicolumn{1}{c|}{77.0} &
  \multicolumn{1}{c|}{68.3} &
  \multicolumn{1}{c|}{58.9} &
  \multicolumn{1}{c|}{53.9} &
  64.5 &
  \multicolumn{1}{c|}{57.3} &
  \multicolumn{1}{c|}{49.4} &
  \multicolumn{1}{c|}{40.7} &
  \multicolumn{1}{c|}{35.0} &
  45.6 \\
BoxInst &
  weakly &
  Yes &
  \multicolumn{1}{c|}{\textbf{82.0}} &
  \multicolumn{1}{c|}{73.3} &
  \multicolumn{1}{c|}{56.6} &
  \multicolumn{1}{c|}{43.6} &
    63.9 &
  \multicolumn{1}{c|}{76.8} &
  \multicolumn{1}{c|}{67.0} &
  \multicolumn{1}{c|}{57.2} &
  \multicolumn{1}{c|}{34.0} &
   58.8 \\
Bayesian-Amodal &
  weakly &
  Yes &
  \multicolumn{1}{c|}{72.3} &
  \multicolumn{1}{c|}{69.6} &
  \multicolumn{1}{c|}{66.2} &
  \multicolumn{1}{c|}{58.5} &
  66.7 &
  \multicolumn{1}{c|}{65.3} &
  \multicolumn{1}{c|}{65.0} &
  \multicolumn{1}{c|}{64.3} &
  \multicolumn{1}{c|}{\textbf{61.4}} &
  64.0 \\
Bayesian-Amodal &
  weakly &
  No &
  \multicolumn{1}{c|}{69.9} &
  \multicolumn{1}{c|}{68.1} &
  \multicolumn{1}{c|}{63.2} &
  \multicolumn{1}{c|}{47.3} &
  62.1 &
  \multicolumn{1}{c|}{58.3} &
  \multicolumn{1}{c|}{59.8} &
  \multicolumn{1}{c|}{58.6} &
  \multicolumn{1}{c|}{53.5} &
  57.6 \\ \hline
Ours &
  weakly &
  No &
  \multicolumn{1}{c|}{81.6} &
  \multicolumn{1}{c|}{\textbf{74.5}} &
  \multicolumn{1}{c|}{\textbf{73.7}} &
  \multicolumn{1}{c|}{\textbf{63.6}} &
  \textbf{73.4} &
  \multicolumn{1}{c|}{\textbf{80.3}} &
  \multicolumn{1}{c|}{\textbf{76.5}} &
  \multicolumn{1}{c|}{\textbf{69.9}} &
  \multicolumn{1}{c|}{57.9} &
  \textbf{71.2} \\ 
  \bottomrule[1.4pt]
\end{tabular}%
}
\caption{The comparison of amodal segmentation performance on the synthetic-occlusion OccludedVehicles dataset and the real-occlusion KINS and COCOA-cls datasets.}
\label{tab:kins-cocoa}
\end{table*}

\subsection{Implementation Details}
We implement our model based on BoxInst~\cite{tian2021boxinst} and Detectron2~\cite{wu2019detectron2} on the PyTorch framework~\cite{paszke2019pytorch}. We use an FCOS module~\cite{tian2019fcos, tian2020conditional} to detect objects and a ResNet-50-FPN backbone~\cite{he2016deep, lin2017feature} to extract multi-scale features for subsequent processes. In the training, we use data at all occlusion levels and choose $\alpha_1^a=2.0,\alpha_2^a=1.0,\alpha_3^a=1.0$. We conduct 60000 iterations with a batch size of 6, during which we adopt a 3-stage learning rate of 0.01 in the first 40000 iterations, 0.001 in the 40000-54000 iterations, and 0.0001 in the 54000-60000 iterations. The training is completed on 3 NVIDIA GeForce RTX 2080Ti GPUs taking about 5 hours per time. In the testing, we evaluate objects at each occlusion level separately to obtain the mean IoU at each level, and then use their average value as the mean IoU of the entire test set.

\subsection{Comparison with Existing Methods}
\paragraph{Baselines.} We benchmark our proposed BLADE method against BBTP~\cite{hsu2019weakly}, BoxInst~\cite{tian2021boxinst}, and Bayesian-Amodal~\cite{sun2022amodal}. These baselines are all under box-level supervision, among which BBTP and BoxInst are the two best-performing methods for box-level supervised general segmentation, and Bayesian-Amodal is a state-of-the-art approach for box-level supervised amodal segmentation.
For the convenience of practical applications, our method only requires the bounding box $\mathbf{B}_{test}$ of the visible portion as the input, which is supported by our expansion-based path. But other methods may require inputting the bounding box of the entire object to know the object center $c$. Bayesian-Amodal provides multiple options, among which we choose both the end-to-end known-$c$ model and the end-to-end unknown-$c$ model for comparison. Designed for normal segmentation, BBTP and BoxInst can only segment the visible parts without knowing $c$ but cannot predict the amodal masks. To compare their performance of amodal segmentation with our method, we adopt the known-$c$ setting for them.

\paragraph{Synthetic Occlusion.} As shown in Table~\ref{tab:kins-cocoa}, our proposed method significantly outperforms existing methods on the OccludedVehicles dataset. Our model performs best at all occlusion levels. Especially in levels with high occlusion ratios, our method shows remarkable advantages. Compared with Bayesian-Amodal~(unknown $c$), our model improves by more than 12\% in mean IoU.

\paragraph{Real Occlusion.} As shown in Table~\ref{tab:kins-cocoa}, our method also performs well on the KINS dataset which is based on real-world images. Although our method is weakly-supervised, which is not comparable to fully-supervised methods, we provide some results in Table~\ref{tab:kins-cocoa} for better evaluation including SAM~\cite{kirillov2023segment}, VRSP~\cite{xiao2021amodal}, and AISFormer~\cite{tran2022aisformer}. Among all these methods, our approach achieves very competitive mean performance. Compared with Bayesian-Amodal~(unknown $c$), our model improves by over 11\% in mean IoU.

\paragraph{Transferability.} Since the number of objects in the COCOA-cls dataset is quite small, we transfer the model trained on the OccludedVehicles dataset to the COCOA-cls dataset, and use the union of the training set and the test set of COCOA-cls for evaluation. As shown in Table~\ref{tab:kins-cocoa}, our model reflects better transferability than others. Compared with Bayesian-Amodal~(unknown $c$), our model improves by more than 13\% in mean IoU.

\begin{figure}[ht]
\centering
\includegraphics[width=0.82\columnwidth]{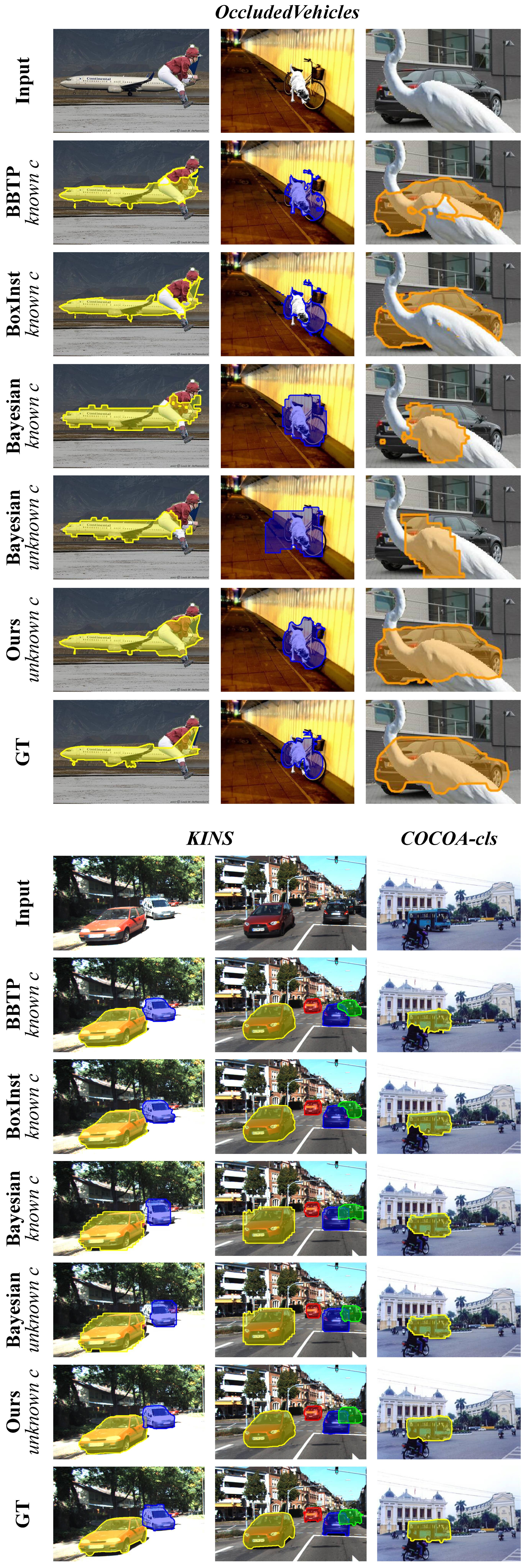}
\caption{Qualitative examples of our approach compared with corresponding ground-truth amodal masks and estimations of BBTP, BoxInst, Bayesian-Amodal~(both the known-$c$ model and the unknown-$c$ model). Zoom in for a better view.}
\label{comparison}
\end{figure}

Some qualitative results are also shown in Fig.~\ref{comparison}, which indicate that masks predicted by our method possess higher resolution and accuracy than others. Our predictions cover more complete occluded segments than BBTP and BoxInst~(such as the car in the 3rd column), while also exhibiting more precise and smooth boundaries than Bayesian-Amodal~(such as the aeroplane in the 1st column).

\subsection{Ablation Study}
We conduct the ablation study on the KINS dataset, the result of which is shown in Table~\ref{tab:ablation}. In these experiments, we all adopt the unknown $c$ setting.

\paragraph{The Effect of Fusion Structure.} To verify the effectiveness of the fusion structure, we cancel the region-branch and the structure of fusing multiple branches' results in the baselines from the 4th row to the 8th row, directly taking the prediction of the amodal-branch as output. Compared with these experimental results, the corresponding results with the fusion structure show significantly better performance.

\paragraph{The Effect of Neighbor Loss.} To validate the importance of the neighbor loss, we conduct the experiments at the 3rd and 4th rows. Compared with the results at the 1st and 2nd rows, their performance show a notable drop, especially at levels with high occlusion ratios.

\paragraph{The Effect of Uniform Loss.} To evaluate the effect of the uniform loss, we introduce the baselines at the 2nd and 4th rows which have no uniform loss. Consequently, a gap emerges between the performance of them and the performance of corresponding models with the uniform loss.

\paragraph{The Effect of Different Weights in $L^a$.} Small adjustments of the weights in $L^a$ will result in certain but not dramatic performance changes, and our selected weights $\alpha_1^a=2.0,\alpha_2^a=1.0,\alpha_3^a=1.0$ achieve good performance.

\section{Conclusion}
In this work, we achieve box-level supervised amodal segmentation through directed expansion. Our approach introduces a fusion structure based on the overlapping region. Conservative strategy and expansion-encouraged strategy are applied to non-overlapping regions and overlapping regions, respectively. Utilizing predicted visible masks as clues, a connectivity loss is incorporated for reasonable expansion. Experimental results indicate our method significantly outperforms other state-of-the-art methods.

\begin{table}[ht]
\centering
\resizebox{\columnwidth}{!}{%
\begin{tabular}{c|c|c|c|c|c|c|c|c}
\toprule[1.4pt]
 &
  \begin{tabular}[c]{@{}c@{}}UN\end{tabular} &
  \begin{tabular}[c]{@{}c@{}}NE\end{tabular} &
  FS &
  FG-0 &
  FG-1 &
  FG-2 &
  FG-3 &
  Mean \\ \hline
1 &
  \checkmark &
  \checkmark &
  \checkmark &
  81.6 &
  74.5 &
  \textbf{73.7} &
  \textbf{63.6} &
  \textbf{73.4} \\
2 &            & \checkmark & \checkmark & 81.6          & \textbf{74.9}          & 70.2 & 57.3 & 71.0   \\
3 & \checkmark &            & \checkmark & 82.8          & 73.2 & 56.8 & 41.6 & 63.6 \\
4 &            &            & \checkmark & \textbf{82.9}          & 72.8          & 56.7 & 40.3 & 63.2 \\
5 & \checkmark & \checkmark &            & 76.6          & 66.7          & 63.9 & 56.2 & 65.9 \\
6 &            & \checkmark &            & 77.3          & 66.9          & 62.9 & 53.2 & 65.1 \\
7 & \checkmark &            &            & 82.3 & 74.2          & 60.0 & 44.7 & 65.3 \\
8 &            &            &            & 82.2          & 73.1          & 56.2 & 40.0 & 62.9 \\ 
\bottomrule[1.4pt]
& $\alpha_1^a$ & $\alpha_2^a$ & $\alpha_3^a$ & FG-0 & FG-1 & FG-2 & FG-3 & Mean \\ \hline
a & 1.0   & 1.0   & 1.0   & 81.3 & 72.3 & 69.6 & 62.1 & 71.3 \\
b & \textbf{2.0}   & \textbf{1.0}   & \textbf{1.0}   & 81.6 & 74.5 & 73.7 & 63.6 & \textbf{73.4} \\
c & 1.0   & 2.0   & 1.0   & 82.0 & 73.4 & 72.7 & 64.8 & 73.2 \\
d & 1.0   & 1.0   & 2.0   & 79.9 & 71.6 & 68.3 & 60.1 & 70.0 \\ 
\bottomrule[1.4pt]
\end{tabular}%
}
\caption{The ablation study on the KINS dataset. UN, NE, and FS respectively represent the uniform loss, the neighbor loss, and the fusion structure. The fusion structure indicates whether to adopt $L^r$ and the corresponding region-branch.}
\label{tab:ablation}
\end{table}

\clearpage

\section{Acknowledgments}
This work was partially supported by the Natural Science Foundation of China under contract 62088102. This work was also partially supported by Qualcomm. We also acknowledge High-Performance Computing Platform of Peking University for providing computational resources.

\bibliography{aaai24}

\end{document}